\documentclass[letterpaper]{article} 
\usepackage{aaai2026}                 
\usepackage{times}                    
\usepackage{helvet}                   
\usepackage{courier}                  
\usepackage[hyphens]{url}             
\usepackage{graphicx}                 
\usepackage{natbib}                   
\usepackage{caption}                  
\usepackage{amsmath}
\usepackage{amsfonts}
\usepackage{mathrsfs}
\usepackage{booktabs}
\usepackage{multirow}
\usepackage[table]{xcolor}
\usepackage{colortbl}
\usepackage{newfloat}
\usepackage{listings}
\usepackage{graphicx}
\usepackage{algorithm}
\usepackage{algorithmicx} 
\usepackage{algcompatible}
\usepackage{bibentry}
\nocopyright
\usepackage{graphicx}
\algrenewcommand\algorithmicrequire{\textbf{Require}}
\algrenewcommand\algorithmicensure{\textbf{Ensure}}

\DeclareCaptionStyle{ruled}{labelfont=normalfont,labelsep=colon,strut=off} 
\floatstyle{ruled}
\newfloat{listing}{tb}{lst}{}
\floatname{listing}{Listing}
\title{Debiasing Diffusion Priors via 3D Attention for Consistent Gaussian Splatting}

\author{
    Shilong Jin\textsuperscript{\rm 1}, Haoran Duan\textsuperscript{\rm 2}, Litao Hua\textsuperscript{\rm 1}, Wentao Huang\textsuperscript{\rm 1}, Yuan Zhou\textsuperscript{\rm 1}\thanks{Corresponding author.}
}

\affiliations{
    \textsuperscript{\rm 1}Nanjing University of Information Science and Technology\\
    \textsuperscript{\rm 2}Tsinghua University\\
    202312490769@nuist.edu.cn, haoran.duan@ieee.org, \{202412621441, 202412491490, zhouyuan\}@nuist.edu.cn
}

\begin{document}

\maketitle

\begin{abstract}
Versatile 3D tasks (e.g., generation or editing) that distill from Text-to-Image (T2I) diffusion models have attracted significant research interest for not relying on extensive 3D training data. However, T2I models exhibit limitations resulting from prior view bias, which produces conflicting appearances between different views of an object. This bias causes subject-words to preferentially activate prior view features during cross-attention (CA) computation, regardless of the target view condition. To overcome this limitation, we conduct a comprehensive mathematical analysis to reveal the root cause of the prior view bias in T2I models. Moreover, we find different UNet layers show different effects of prior view in CA. Therefore, we propose a novel framework, \textit{\textbf{TD-Attn}}, which addresses multi-view inconsistency via two key components: (1) the 3D-Aware Attention Guidance Module~{(\textbf{3D-AAG})} constructs a view-consistent 3D attention Gaussian for subject-words to enforce spatial consistency across attention-focused regions, thereby compensating for the limited spatial information in 2D individual view CA maps; (2) the Hierarchical Attention Modulation Module (\textbf{HAM}) utilizes a Semantic Guidance Tree (SGT) to direct the Semantic Response Profiler (SRP) in localizing and modulating CA layers that are highly responsive to view conditions, where the enhanced CA maps further support the construction of more consistent 3D attention Gaussians. Notably, HAM facilitates semantic-specific interventions, enabling controllable and precise 3D editing. Extensive experiments firmly establish that TD-Attn has the potential to serve as a universal plugin, significantly enhancing multi-view consistency across 3D tasks.
\end{abstract}


\section{Introduction}

{Text-driven 3D generation \cite{ding2024text,hong2022avatarclip,metzer2023latent,michel2022text2mesh,guo2023stabledreamertamingnoisyscore} and editing \cite{wu2024gaussctrl,koo2024posterior,Li_Dou_Shi_Lei_Chen_Zhang_Zhou_Ni_2024} are increasingly applied in industrial design, VR/AR, and digital content production. These technologies allow the creation and modification of 3D assets based on natural language input. Constrained by the complexity of 3D representation learning and the scarcity of 3D data, current text-driven 3D technologies \cite{liang2024luciddreamer,li2024connecting,li2024dreamscene,di2025hyper} typically combine 3D Gaussian Splatting (3DGS) \cite{3dgs} representation with distillation from text-to-image (T2I) models. Specifically, training-free distillation techniques eliminate the dependency on 3D data, while the differentiable 3DGS representation replaces Neural Radiance Fields (NeRF) \cite{mildenhall2021nerf} by enabling high-quality rendering with significant acceleration during inference.}
\begin{figure}[t]
\centering
\includegraphics[width=1\columnwidth]{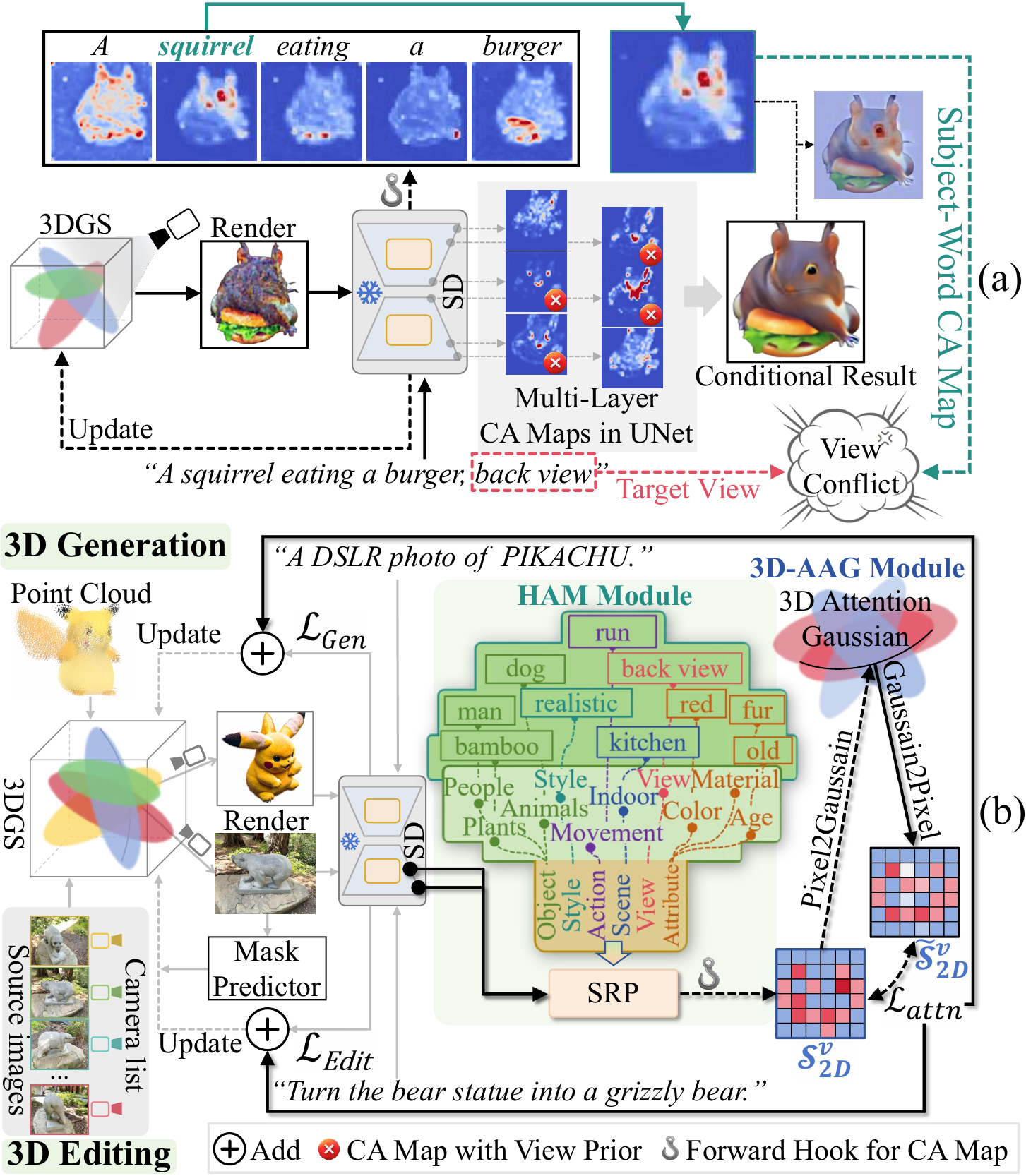} 
\caption{(a) Subject-word CA map anomalies corrupt denoising outputs with layer-varying bias intensity. (b) Overview of our TD-Attn framework integrating HAM and 3D-AAG modules for consistent 3D generation and editing.}
\label{fig1}
\end{figure}

However, a fundamental and pervasive challenge persists across existing 3D generation and editing frameworks: the inability to guarantee multi-view consistency. This manifests most prominently as the Janus problem \cite{janus,hong2023debiasing,poole2022dreamfusion}, where 3D objects have conflicting faces or orientations when rendered from different angles. This issue significantly undermines the realism and coherence of the text-to-3D output. Therefore, ensuring view consistency in 3D tasks is crucial for high-quality results, which is the motivation for this paper.

Previous studies \cite{janus,zhang2024viewpoint,liu2023zero,huang2024dreamcontrol} attribute this inconsistency to the view bias of T2I diffusion models, which arises from training datasets with skewed view distributions and sparse view annotations. We theoretically validate and visualize the impact of prior view bias, as shown in Fig.~\ref{fig1}(a). Under a ``back view" condition, we input the prompt token with view annotations and its corresponding view-rendered result into the frozen UNet \cite{ronneberger2015u}. The token-level attention maps show that the subject-word token (e.g., ``squirrel") activates frontal-view features, and the spatial distribution of these activations correlates with the distortions observed in the conditional result. This phenomenon demonstrates that the imbalanced training data of T2I models fundamentally shapes the inductive biases of subject-word token cross-attention (CA) scores, which represent the attention regions of words in prompts across spatial features of noisy images. Since high attention scores correspond to regions where generation and editing are concentrated, this bias directly affects the spatial distribution of generated content. Specifically, the abundance of training samples without view annotations paired with prior view images causes subject-words to have higher similarity scores with prior view images, making prior view features activate more easily. When processing prompts without explicit view cues, the network defaults to generating prior view content due to these learned associations. Even when processing prompts with explicit view cues, this prior view preference dilutes or overrides the view conditions, leading to the Janus problem in 3D tasks. Additionally, Fig.~\ref{fig1}(a) shows CA maps across multiple UNet layers, which highlights the layer-wise heterogeneity \cite{feng2023efficient} of prior view preference within the UNet.

To address this prior view preference-dominated inconsistency, we first provide a mathematical analysis of how prior view preference in T2I models compromises 3D generation consistency. The analysis reveals that the inherent view bias in T2I models causes conflicts between subject-word and view control conditions, resulting in significant negative gradients during the 3D generation process. Therefore, we propose TD-Attn, a novel framework shown in Fig.~\ref{fig1}(b), which comprises two modules: the 3D-Aware Attention Guidance Module (3D-AAG) and the Hierarchical Attention Modulation Module (HAM). First, 3D-AAG constructs a 3D attention Gaussian for subject-word tokens significantly influenced by view preference, which accumulates multi-view 2D attention maps through inverse rendering. The Gaussian effectively dilutes erroneous attention patterns dominated by prior view preference and compensates for the limited spatial information in CA maps from independent 2D denoising processes.

Furthermore, to address the prior view preference differences across different UNet layers, we construct a hierarchical Semantic Guidance Tree (SGT) containing rich semantic words with the assistance of large language models (LLMs). HAM then utilizes the SGT to guide a Semantic Response Profiler (SRP) to localize UNet layers and CA heads with high-response view semantics, and modulates these CA layers to obtain multi-view consistent CA maps, which further support the construction of the 3D attention Gaussian. In particular, HAM not only tracks view semantics, but also enables effective localization and control of other semantics, such as color and material, providing fine-grained control over target semantics in 3D editing tasks. In summary, this paper makes the following contributions:
\begin{itemize}
\item We provide a comprehensive mathematical analysis demonstrating how prior view preference affects 3D tasks, with CA mechanism analysis serving as visual validation of this principle. The CA maps reveal that prior view preference from subject-word tokens override the view information from view description tokens, leading us to formulate an optimization objective centered on suppressing prior view preference in subject-word tokens.
\item We propose 3D-AAG, which accumulates 2D CA maps from different views to construct a 3D attention Gaussian for subject-word tokens. 3D-AAG then constrains anomalous 2D CA scores to guarantee multi-view consistency.
\item We propose HAM, which uses the semantic guidance tree to guide SRP in tracking and weighting CA layers that are highly responsive to view semantics. Then view-enhanced CA maps are further used to construct a more consistent 3D attention Gaussian.
\item Proposed HAM enables targeted, semantic-specific interventions during 3D editing, giving precise control over semantics such as color, and material.
\end{itemize}

\section{Related Work}
\subsection{3D Gaussian Splatting}
3D Gaussian Splatting is an explicit and efficient 3D scene representation that models environments using anisotropic 3D Gaussians. Each Gaussian is characterized by its 3D position $\mu \in \mathbb{R}^3$, covariance matrix $\boldsymbol{\Sigma} \in \mathbb{R}^{3 \times 3}$, opacity $\eta$, and view-dependent color $c$ represented through spherical harmonics coefficients. The representation enables real-time, differentiable rendering \cite{ai1Huang_Wu_Deng_Gao_Gu_Liu_2025,ai2Lin_Li_Huang_Tang_Liu_Liu_Wu_Song_Yang_2025,ai3Li_Lv_Yang_Huang_2025,ai5Zhou_Fan_Chen_Huang_Liu_Li_2025,ai6Qu_Li_Cheng_Shi_Meng_Ma_Wang_Deng_Zhang_2025}, through a splatting technique that projects these 3D Gaussians onto 2D image planes while preserving their geometric properties. During optimization, the covariance matrix is factorized into rotation $\boldsymbol{R}$ and scaling $\boldsymbol{S}$ components to ensure numerical stability. A key advantage of 3D Gaussian Splatting is its adaptive densification strategy, where Gaussians dynamically split or clone during training to capture fine scene details while preserving computational efficiency. 
\subsection{Text-driven 3D Generation}
Early text-to-3D generation methods \cite{jain2022zero} leverage the cross-modal capabilities of CLIP \cite{hong2022avatarclip} but suffer from limited alignment precision. The emergence of diffusion models \cite{rombach2022high} establishes a new paradigm \cite{poole2022dreamfusion,tang2023dreamgaussian} where 2D diffusion priors guide differentiable 3D representations through distillation techniques, significantly improving generation quality. However, existing methods still face view inconsistency issues where generated 3D objects exhibit the multi-face Janus problem. While some approaches \cite{shi2023mvdream,xie2024carve3d,hong2023debiasing} fine-tune diffusion models with multi-view datasets or employ geometric constraints and prompt engineering, they either require additional training overhead or lack precise view control. These limitations underscore the need for robust solutions that ensure multi-view consistency without extensive retraining.
\begin{figure}[t]
\centering
\includegraphics[width=0.95\columnwidth]{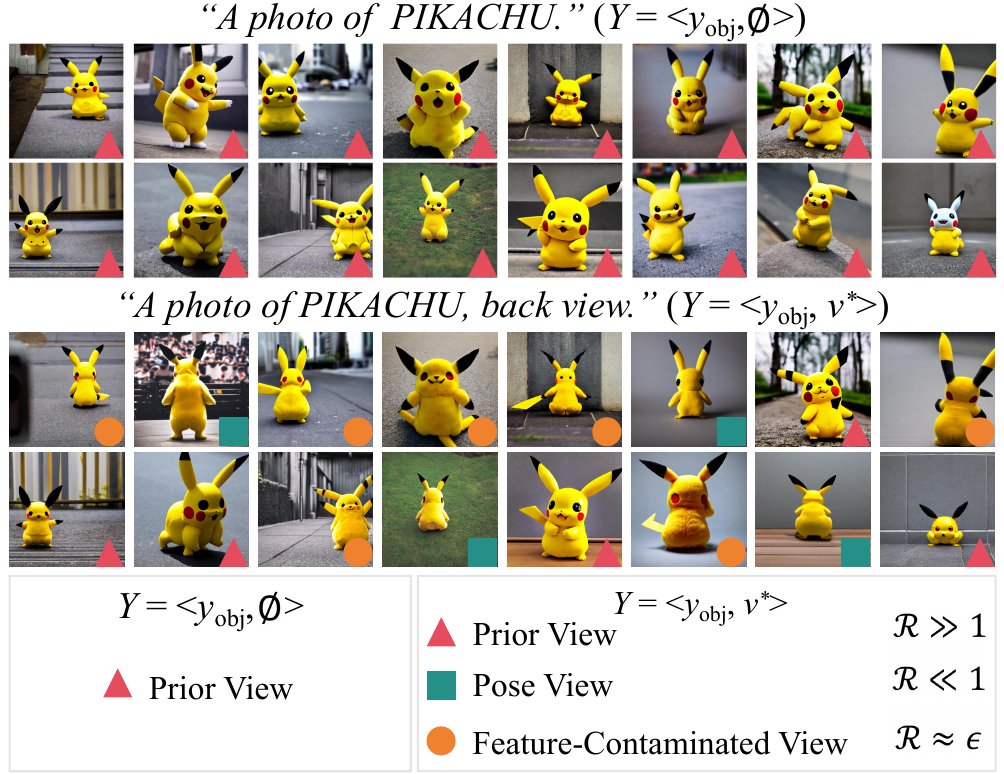} 
\caption{T2I model generation results demonstrating view bias effects. Top: results without view annotations show predominant prior view generation. Bottom: results with ``back view" condition exhibit three phenomena: Prior View (red), successful Target View (green), and Feature-Contaminated View (orange).}
\label{fig3}
\end{figure}
\begin{figure*}
\centering
\includegraphics[width=1\textwidth]{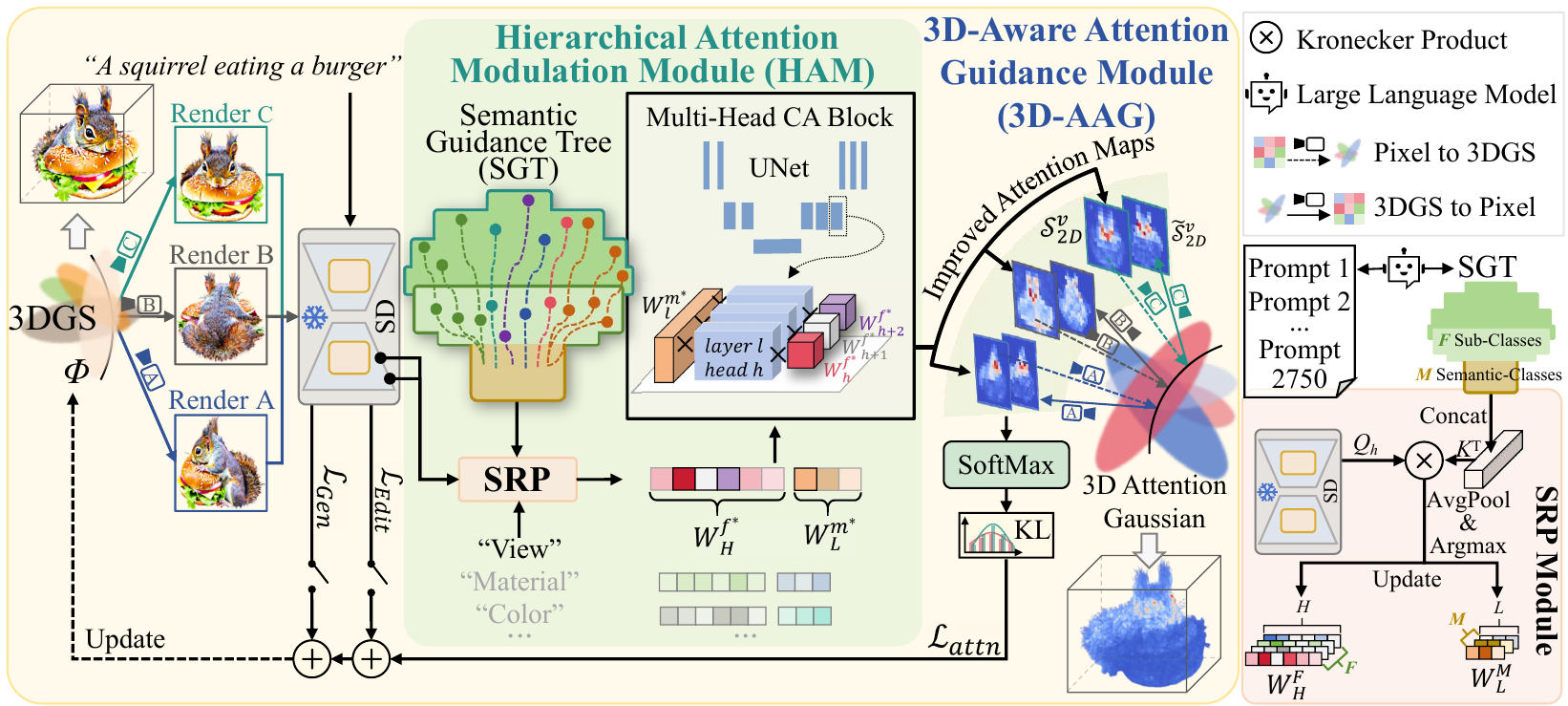} 
\caption{Overview of the TD-Attn framework. HAM guides the CA layers in UNet towards target semantics by weighting them through SRP, which is directed by the SGT. The view enhanced CA maps of subject-word token are then inversely mapped to 3DGS to generate a 3D attention Gaussian with spatial consistency. Finally, the newly rendered subject-word CA maps from the 3D attention Gaussian are used to suppress the prior preference information in the original attention maps.}
\label{fig2}
\end{figure*}

\section{Method}
In this section, we present our innovative framework, TD-Attn, which is developed based on the theoretical analysis of prior view bias in the first section. The pipeline of TD-Attn is illustrated in Fig.~\ref{fig2}. In the subsequent sections, we explore the two primary components of our methodology, which effectively guarantee view consistency in 3D tasks.

\subsection{Theoretical Analysis of View Inconsistency}
In T2I diffusion models, image generation can be formulated as the conditional probability distribution $p_\theta(x|Y)$, where $\theta$ represents the parameters of the model that govern the generation process, and $x$ represents the generated image given a text prompt $Y$. The text prompt ${Y = \langle y_{\text{obj}}, v^*\rangle}$ consists of object description components $y_{\text{obj}}$ and view conditions $v^*$. The training dataset ${\mathcal{D} = \{\langle x_i, Y_i\rangle \}_{i=1}^\mathcal{N}}$ typically exhibits significant view distribution bias and sparse view annotations, which can be formalized as: ${p_\mathcal{D}({v_{\text{prior}}}|y_{\text{obj}}) \gg p_\mathcal{D}({v_{\text{other}}}|y_{\text{obj}})}$ and ${p_\mathcal{D}(Y_{v^*} = \emptyset) \gg p_\mathcal{D}(Y_{v^*} \neq \emptyset)}$. 
During training, when the model learns from text prompts without view annotations, i.e. ${Y = \langle y_{\text{obj}}}, \emptyset\rangle$, the view information solely derives from the prior view distribution, i.e. ${p(v|Y) = p(v|y_{\text{obj}},\emptyset) \propto p_\mathcal{D}(v|y_{\text{obj}})}$, this reveals that when processing text without view annotations, T2I models implicitly tend to favor prior view preference generation as the optimal solution. As shown in the upper part of Fig.~\ref{fig3}, when generating from prompts such as ``A photo of PIKACHU", the model generates images based on the prior view preference.

When processing prompts containing view conditions, the ideal view prior should be $p(v|Y) = \delta_{v,v^*}$, where $\delta_{v,v^*}$ is the Dirac delta function, and the model's training objective is $p_\theta^{\text{ideal}}(x|Y) = p_\theta(x|y_{\text{obj}}, v^*)$. However, the learned view distribution deviates from the ideal Dirac distribution due to training data bias, causing subject-word tokens to respond strongly to prior view features. We model this deviation using a bias coefficient $\epsilon$ where $0 < \epsilon \ll 1$:
\begin{equation}
p(v|Y) = (1-\epsilon) \delta_{v,v^*} + \epsilon p_\mathcal{D}(v|y_{\text{obj}}),
\label{2d0}
\end{equation}
we then analyze the probability ratio by substituting $v_{\text{prior}}$ and $v^*$ into $p(v|Y)$:
\begin{equation}
\mathcal{R} = \frac{p(v_{\text{prior}}|Y)}{p(v^*|Y)} = \frac{\epsilon p_\mathcal{D}(v_{\text{prior}}|y_{\text{obj}})}{(1-\epsilon) + \epsilon p_\mathcal{D}(v^*|y_{\text{obj}})},
\label{2d1}
\end{equation}
Eq.~(\ref{2d0}) identifies three distinct generation phenomena: When $\mathcal{R} \ll 1$, the target view $v^*$ dominates, enabling successful ``back view'' generation (green square marks in Fig.~\ref{fig3}); when $\mathcal{R} \gg 1$, prior view bias overwhelms the target condition, resulting in frontal Pikachu generation despite the ``back view'' condition (red triangle marks); when $\mathcal{R} \approx \epsilon$, comparable view probabilities cause feature contamination, where the model simultaneously responds to both prior and target view conditions, producing images with mixed characteristics such as back-view poses with frontal facial features (orange circle marks).

In 3D tasks, these anomalous CA scores between subject-word tokens and prior view features accumulate across the multi-view optimization process. Specifically, the model optimizes a 3D representation $\phi$ by leveraging multiple 2D supervision signals from different views. It can be formulated as :
\begin{equation}
\tilde{P}_{3D}(\phi) = \prod_{v^* \in \Lambda} p_{2D}(z_\phi|y_{\text{obj}}, v^*),
\end{equation}
where $\Lambda$ represents the set of camera views, and $z_\phi$ denotes the rendered 2D image from the 3D representation $\phi$ at view $v^*$.
The gradient with respect to the 3D parameters is:
\begin{equation}
\begin{aligned}
\nabla_{\phi} \log p_{3D}(\phi)& 
\\= |\Lambda| \mathbb{E}_{v^* \in \Lambda}&[\nabla_{z_\phi} \log p_{2D}(z_\phi|v^*, y_{\text{obj}}) \frac{\partial z_\phi}{\partial \phi}] \\
= |\Lambda| \mathbb{E}_{v^* \in \Lambda}&[(\nabla_{z_\phi} \log p_{2D}(z_\phi) \\&+ \nabla_{z_\phi} \log p_{2D}(v^*, y_{\text{obj}}|z_\phi)) \frac{\partial z_\phi}{\partial \phi}],
\end{aligned}
\end{equation}
to focus on the influence of view priors bias in generation, we can safely neglect the unconditional term $\nabla_{z_\phi} \log p_{2D}(z_\phi)$. While the unconditional term captures the global structure of the generated image, its effect on local features such as view is minimal in this context. Consequently, we further decompose the conditional term:
\begin{equation}
\begin{aligned}
\nabla_{z_\phi} \log p_{2D}(v^*, y_{\text{obj}}|z_\phi) = \nabla_{z_\phi} \log p_{2D}(v^*|z_\phi)\\
+ \nabla_{z_\phi} \log p_{2D}(y_{\text{obj}}|z_\phi) + \nabla_{z_\phi} \log C,
\end{aligned}
\label{2d2}
\end{equation}
where $C = \frac{p_{2D}(v^*, y_{\text{obj}}|z_\phi)}{p_{2D}(v^*|z_\phi)p_{2D}(y_{\text{obj}}|z_\phi)} = \frac{p_{2D}(v^*|y_{\text{obj}}, z_\phi)}{p_{2D}(v^*|z_\phi)}$. Based on our previous analysis, when certain object descriptions become overly common under prior views, strong dependencies emerge between $y_{\text{obj}}$ and $v_\text{prior}$. In this case, when $v^*$ approaches $v_\text{prior}$, i.e, $C \rightarrow 1$, the term $\nabla_{z_\phi} \log C$ does not significantly interfere with gradient computation. However, when $v^*$ is far from $v_\text{prior}$, we have $p_{2D}(v^*|y_{\text{obj}}, z_\phi) \ll p_{2D}(v^*|z_\phi)$, resulting in $C \rightarrow 0$. Consequently, $\nabla_{z_\phi} \log C \ll 0$, resulting in large negative gradient effects that significantly impact the generation process.

Eq.~\eqref{2d2} directly explains the empirical observations in Fig.~\ref{fig1}(a), where the large negative gradients from the $\nabla_{z_\phi} \log C$ term correspond to the anomalous attention focus regions we observe in practice. This mathematical correspondence reveals why subject-word tokens consistently activate frontal-view features under ``back view" conditions, creating the distortions visible in the conditional results. To mitigate this issue, we suppress the prior view preference in subject-word token attention, as shown in Fig.~\ref{fig2}.

\subsection{3D-Aware Attention Guidance Module (3D-AAG)}
During each denoising step through CA layers in UNet, 2D CA maps are inversely mapped back to the 3DGS representation \cite{chen2024gaussianeditor} to accumulate multi-view 2D attention weights. This enables us to construct a view-consistent 3D attention Gaussian, which compensates for the limited spatial information in CA scores computed from independent views. Thanks to the explicit nature of 3DGS, the accumulated attention weight for each Gaussian $i$ can be computed as:
\begin{equation}
w_i = \sum_{v \in \Lambda} \sum_{p \in \mathcal{I}(\mathcal{S}_{2D}^v)} [o_i(p)  T_i^v(p)  \mathcal{I}(\mathcal{S}(p)_{2D}^v)],
\end{equation}
where $o_i(p)$ is the opacity of Gaussian~$i$, $T_i^v(p)$ denotes the transmittance along the ray from pixel $p$ to Gaussian $i$ at view $v$, $\mathcal{I}(\mathcal{S}(p)_{2D}^v)$ is the interpolated attention score of the subject-word token at pixel $p$, and $\mathcal{S}_{2D}^v$ denotes the CA map of the subject-word token at view $v$ defined as:
\begin{equation}
\mathcal{S}_{2D}^v = \text{Softmax}\left(\frac{Q_v K_{\text{sbj}}^T}{\sqrt{d}}\right),
\end{equation}
where $Q_v$ is the query matrix projected from the spatial features of the noisy image with dimension $d$ at view $v$ and $K_{\text{sbj}}$ is the key corresponding to the subject-word token. Since the 2D attention maps have different resolutions from the rendered images, bilinear interpolation $\mathcal{I}(\cdot)$ is applied to match $\mathcal{S}_{2D}^v$ to the rendering resolution, which ensures that each Gaussian accumulates attention weights from multiple views, effectively diluting the prior view bias and promoting view-consistent 3D parameter updates.

During 3D optimization, 3DGS performs adaptive densification and pruning \cite{3dgs} to adjust Gaussian distributions. The 3D attention Gaussian synchronizes with these operations, which ensures view-consistent attention patterns throughout optimization while maintaining alignment with the evolving geometric representation.

Since the 3D attention Gaussian integrates CA information from multiple views, it contains more accurate view information compared to any individual 2D CA map obtained from a single optimization step. Therefore, the accumulated 3D attention Gaussian acquires the capability to guide 2D attention maps. Specifically, to emphasize the focus regions of the rendered attention map $\widetilde{\mathcal{S}}_{2D}^v$ at view $v$, we apply softmax normalization and employ KL divergence loss \cite{kingma2014semi} to constrain the discrete 2D CA maps. The attention loss $\mathcal{L}_{\text{attn}}$ can be formulated as:
\begin{equation}
\mathcal{L}_{\text{attn}} = \text{KL}\left(\text{Softmax}(\widetilde{\mathcal{S}}_{2D}^v) \parallel \mathcal{I}(\mathcal{S}_{2D}^v)\right).
\end{equation}
\subsection{Hierarchical Attention Modulation Module (HAM)}

Eq.~\eqref{2d2} demonstrates that 3D tasks suffer from gradient interference due to prior view preference when distilling T2I models, where the conditional gradient term is compromised by anomalous CA scores between subject-word tokens and prior view features. Therefore, we aim to enhance target view semantics in UNet CA layers to minimize the preference ratio $\mathcal{R}$ in Eq.~\eqref{2d1}, ensuring $\mathcal{R} \ll 1$ and guaranteeing that the posterior probability $p(v|Y)$ in Eq.~\eqref{2d0} conforms to the target view $v^*$. 

We adopt a multimodal perspective to examine the interaction patterns between subject-word token keys and image queries within the CA mechanism. As shown in the bottom right of Fig.~\ref{fig2}, inspired by HRV \cite{park2024cross}, SRP concatenates enriched concept words to form a key matrix $K^T$, which is then matched against extensive image query matrices $Q$ to identify the highest CA scores among semantic words, thereby defining the primary conceptual focus of each attention head. Unlike HRV, we consider the inherent polysemy in natural language, where concept words often carry multiple semantic meanings. For instance, ``match" can refer to a football match, a perfect pairing, or a striking tool for lighting, while ``light" can denote physical illumination, object weight, or conceptual simplicity. To mitigate this ambiguity, we construct a semantic guidance tree using LLMs \cite{brown2020language}. As depicted in Fig.~\ref{fig1}(b), the SGT employs a three-level hierarchical structure: the root level encompasses $M$ distinct Semantic-Classes (e.g., Object, Attribute); the intermediate level contains $F$ Sub-Classes that belong to each Semantic-Class; and the leaf level comprises $F$ specific Instances, randomly selected from each Sub-Class set. Given the hierarchical relationships between CA heads and UNet layers, we compute separate scores for CA heads and UNet layers to correspond to different semantic granularities in the SGT. To precisely localize layers corresponding to prior view preference, we incorporate view conditions into the majority of these prompts.

For semantic localization, similarity matrices between the obtained query matrix $Q$ and randomly selected instance-words key matrix $K^T$ are computed. For $H$ CA-Heads, the relevance score $W_{h}^{f}$ of $F$ Sub-Classes in head $h$ and subclass $f$ is computed as:
\begin{equation}
W_h^{f} = \text{argmax}_{f} \left( \text{AvgPool}\left(\frac{Q_h K^T}{\sqrt{d}}\right) \right),
\end{equation}
and for $L$ UNet layers, the relevance score $W_{l}^{m}$ of $M$ Semantic-Classes in layer~\( l \) and class~\( m \) is computed as:
\begin{equation}
W_l^{m} = \text{argmax}_{m} \left( \sum_{u \in \mathcal{U}_{m}}\sum_{h \in H_l} \text{AvgPool}\left(\frac{Q_h K_{u}^T}{\sqrt{d}}\right) \right),
\end{equation}
where $\mathcal{U}_{m}$ represents the set of Instances belonging to Semantic-Class~$m$, and $H_l$ denotes the set of attention heads in layer $l$. This produces one-hot relevance scores that are accumulated into layer-wise weights $W_L^M \in \mathbb{R}^{M \times L}$ and head-wise weights $W_H^F \in \mathbb{R}^{F \times H}$.
\begin{figure*}
\centering
\includegraphics[width=1\textwidth]{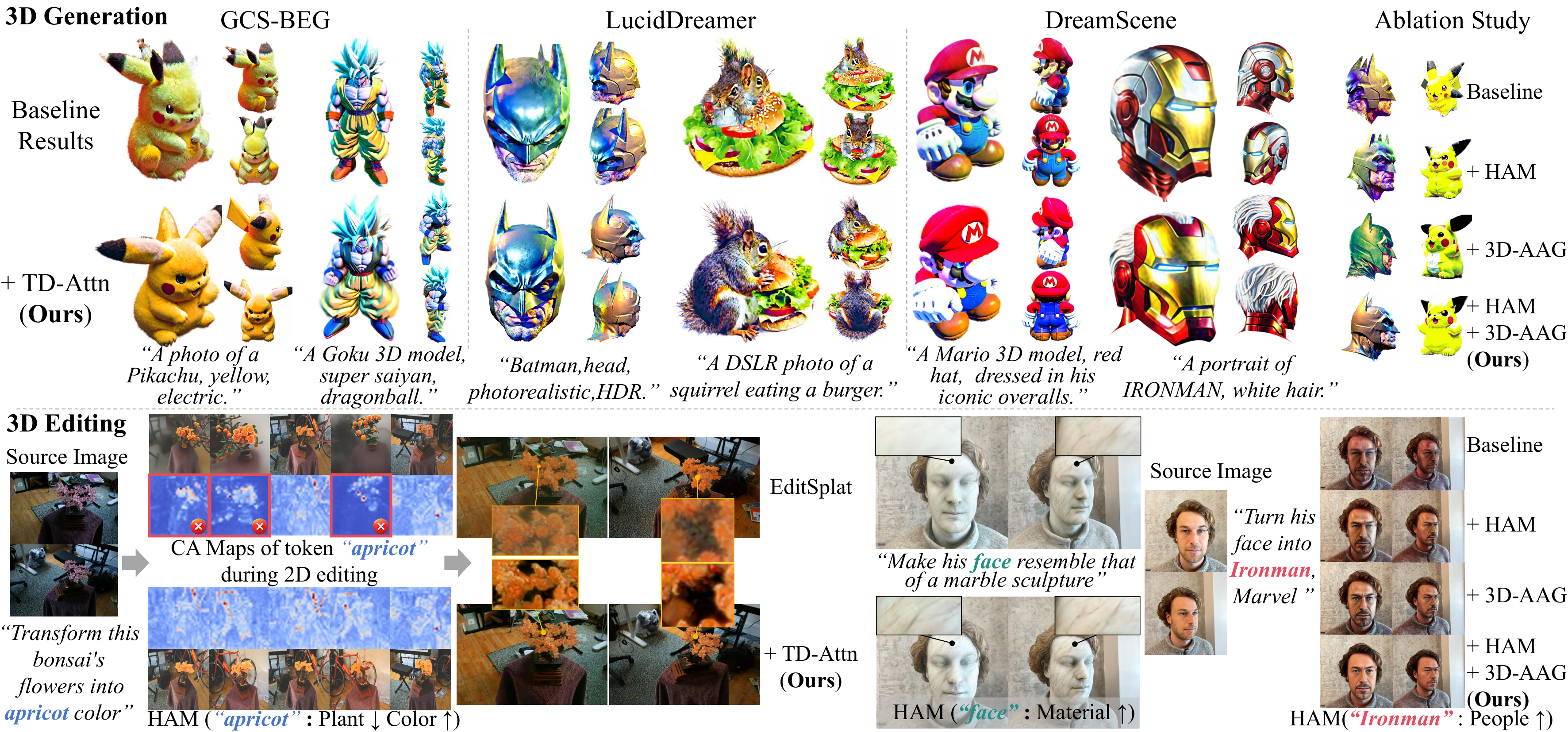} 
\caption{Qualitative comparisons. Top: 3D generation results comparing baseline methods with TD-Attn-enhanced versions, eliminating multi-view inconsistencies and visual artifacts. Right: Ablation study showing progressive improvements. Bottom: 3D editing results demonstrating semantic consistency and multi-view coherence improvements.}
\label{fig4}
\end{figure*}

The HAM then modulates the UNet's attention computation by applying these learned weights for target semantic $f^*$:
\begin{equation}
\hat{\mathcal{A}}_h = \lambda W_l^{m^*}  W_h^{f^*}   \mathcal{A}_h,
\end{equation}
\begin{table*}[t]
\centering
\small
\begin{tabular}{lccccccc}
\toprule
\multirow{2}{*}{\textbf{Methods}} & \multicolumn{2}{c}{\textbf{User Study}} & \multirow{2}{*}{\textbf{ImageReward} $\uparrow$} & \multirow{2}{*}{\textbf{CLIP$_{\text{sim}}$}$\uparrow$} & \multicolumn{2}{c}{\textbf{Frequency of Inconsistency}} \\
\cmidrule(lr){2-3} \cmidrule(lr){6-7}
& \textbf{Quality} $\uparrow$& \textbf{Consistency} $\uparrow$& & & \textbf{f$_{\text{mf}}$(\%)} $\downarrow$& {\textbf{f$_\text{inc}$(\%)}} $\downarrow$\\
\midrule
DreamScene & 4.52 & 3.51 & -0.725 & 0.299 & 46.7 & 66.7 \\
\rowcolor{gray!15}DreamScene + Ours & \textbf{6.26} & \textbf{5.44 }& \textbf{-0.229} & \textbf{0.306} & \textbf{33.3} & \textbf{53.3} \\
LucidDreamer & 5.34 & 4.02 & -0.386 & 0.309 & 26.7 & 60.0 \\
\rowcolor{gray!15}LucidDreamer + Ours & \textbf{7.27} & \textbf{6.67} & \textbf{0.124} & \textbf{0.320 }& \textbf{13.3 }& \textbf{33.3 }\\
GCS-BEG & 6.13 & 4.18 & 0.158 & 0.312 & 33.3 & 60.0 \\
\rowcolor{gray!15}GCS-BEG + Ours & \textbf{7.81} & \textbf{7.68} & \textbf{0.397} & \textbf{0.317} & \textbf{6.7} & \textbf{26.7} \\
\bottomrule
\end{tabular}
\caption{Quantitative comparison on 3D generation task.}
\label{tab: tab1}
\end{table*}
where $f^* \in m^*$ and $h \in H_l$, $\lambda$ is the modulation coefficient, $W_l^{m^*}$ represents the layer-wise weight for Semantic-Class $m^*$ at layer $l$, $W_h^{f^*}$ represents the head-wise weight for Sub-Class $f^*$ at head $h$, $\mathcal{A}_h$ is the original attention computation at head $h$, and $\hat{\mathcal{A}}_h$ is the modulated attention computation.

Through this semantic-aware weighting mechanism, the model's sensitivity to target semantics can be selectively enhanced while suppressing prior biases. Notably, the view-enhanced CA maps produced by HAM further support the construction of the more consistent 3D attention Gaussian, and the improved attention consistency helps $\mathcal{L}_{\text{attn}}$ more effectively suppress prior view preference, creating a synergistic effect between our two proposed components that jointly ensures multi-view consistency in 3D tasks. 
For different 3D tasks, we employ task-specific loss combinations:
\begin{align}
\mathcal{L} &= \mathcal{L}_{\text{Gen}} + \lambda_1 \mathcal{L}_{\text{attn}} \quad \text{for generation tasks}, \\
\mathcal{L} &=  \mathcal{L}_{\text{Edit}} + \lambda_2 \mathcal{L}_{\text{attn}} \quad \text{for editing tasks},
\end{align}
where $\lambda_1$ and $\lambda_2$ are attention balance coefficients.

\section{Experiments}

In this section, we provide a systematic evaluation of TD-Attn through a dual approach integrating qualitative visual analysis and quantitative benchmarking. First, we conduct qualitative comparisons in the first section, followed by comprehensive quantitative analysis in the second section. Finally, we provide ablation experiments to validate the effectiveness of each module in our framework. All experiments were conducted using a single NVIDIA RTX 4090 GPU. Additionally, we employ Stable Diffusion \cite{rombach2022high} v2.1 base model for distillation in generation tasks, and adopt InstructPix2Pix \cite{brooks2023instructpix2pix} based on Stable Diffusion v1.4 to obtain editing results in editing tasks. More implementation details can be found in the appendix.

\subsection{Qualitative Comparisons}
As a universal plugin, TD-Attn can be seamlessly integrated into existing 3D generation and editing frameworks. Fig.~\ref{fig4} demonstrates our method's performance across diverse scenarios. For generation tasks, we integrate TD-Attn into baseline methods including DreamScene \cite{li2024dreamscene}, LucidDreamer \cite{liang2024luciddreamer}, and GCS-BEG \cite{li2024connecting}, with results shown in the second row of Fig.~\ref{fig4}. Compared to baseline approaches, our method successfully eliminates extraneous limbs and facial features that plague existing methods, such as additional ears in Pikachu and Batman, or extra faces in Mario, Ironman, and Squirrel. 

For 3D editing tasks, we employ state-of-the-art (SOTA) EditSplat \cite{lee2025editsplat} as our baseline, with improvements demonstrated in the lower portion of Fig.~\ref{fig4}. When processing prompts such as ``Transform this bonsai's flowers into apricot color", baseline methods often misinterpret ``apricot" as a botanical entity rather than a color descriptor. We visualize the CA maps of token ``apricot" during the 2D editing process. As shown in Fig.~\ref{fig4}, the baseline method's attention maps exhibit erroneous activations on plant-related regions (marked with red crosses), indicating that the model incorrectly associates ``apricot" with botanical entities despite the contextual cue ``color". This semantic misinterpretation translates to inconsistent editing results across multiple views. Similarly, when handling ``Turn his face into Ironman, Marvel", editing results exhibit hybrid states combining human facial features with armor characteristics. TD-Attn provides a way to modulate semantic intensity of specific words within prompts. As visualized in the lower-left portion of Fig.~\ref{fig4}, we enhance the ``color" semantics of ``apricot" while suppressing its ``plant" semantics, ensuring semantic consistency of ``apricot" throughout the editing process. The HAM module successfully redirects attention toward color-relevant regions, as demonstrated in the corrected CA maps. In contrast, baseline methods frequently exhibit semantic materialization of ``apricot" entities, which conflicts with the editing intent. Furthermore, baseline results suffer from multi-view editing inconsistencies, exemplified when processing ``Make his face resemble that of a marble sculpture", where different viewpoints exhibit inconsistent facial textures, a phenomenon that TD-Attn successfully mitigates.
\begin{table}[t]
\centering

\small
\begin{tabular}{lccc}
\toprule
\multirow{2}{*}{\textbf{Methods}} & \multirow{2}{*}{\textbf{CLIP$_{\text{sim}}$}$\uparrow$} & \multicolumn{2}{c}{\textbf{Frequency of Inconsistency}} \\
\cmidrule(lr){3-4}
& & \textbf{f$_{\text{mf}}$(\%)} $\downarrow$ & \textbf{f$_{\text{inc}}$(\%)} $\downarrow$ \\
\midrule
Janus issue &\textbf{0.318} & 100.0 & 100.0 \\
Baseline& 0.307 & 35.6 & 62.2 \\
+ HAM & 0.311 & 20.0 & 57.8 \\
+ 3D-AAG & 0.313 & 24.4 & 44.4 \\
\midrule
TD-Attn & 0.314 & \textbf{17.8} & \textbf{37.8} \\
\bottomrule
\end{tabular}
\caption{Ablation study for generation task.}
\label{tab2}
\end{table}
\begin{table}[t]
\centering
\small

\begin{tabular}{lccc}
\toprule
\textbf{Methods} & \textbf{CLIP$_{\text{sim}}$}$\uparrow$ & \textbf{CLIP$_{\text{dir}}$}$\uparrow$ & \textbf{User Study} $\uparrow$ \\
\midrule
Baseline(EditSplat)& 0.253 & 0.101 & 4.18 \\
+ HAM & 0.272 & 0.112 & 5.26 \\
+ 3D-AAG & 0.261 & 0.102 & 4.73 \\
\midrule
TD-Attn& \textbf{0.277} & \textbf{0.114} & \textbf{6.34} \\
\bottomrule
\end{tabular}
\caption{Ablation study for editing task.}
\label{tab3}
\end{table}
\subsection{Quantitative Comparisons}
To intuitively demonstrate TD-Attn's performance, we compare baseline methods with their TD-Attn integrated counterparts. For generation tasks, we employed standard evaluation metrics ImageReward \cite{xu2023imagereward} and $\text{CLIP}_\text{sim}$ \cite{radford2021learning} as objective measures. Additionally, we evaluated the percentage of generated results exhibiting Inconsistency, denoted as f$_\text{inc}$. We extended the traditional Janus problem definition to encompass broader inconsistencies including extra limbs and other implausible artifacts. For comprehensive evaluation, we selected diverse prompts involving objects with distinct front-back differences (e.g., portraits, animals, vehicles). The quantitative results are summarized in Tab.~\ref{tab: tab1}, where f$_\text{mf}$ denotes the frequency of Janus problem. As shown in the table, our method achieves the highest scores across all metrics, with particularly notable improvements in mitigating the Janus issue, where our method outperforms the baseline average by approximately 50\%.

We conducted a user study comparing our method with the SOTA approaches for subjective evaluation. Using a 1-to-10 scale, participants assessed models based on two key criteria: ``Generation Quality" and ``3D Consistency." All samples were randomly selected without cherry-picking. As shown in Tab.~\ref{tab: tab1}, our method achieves the highest average scores across both evaluation metrics. These results demonstrate that 3D content generated by our approach consistently exhibits superior visual appeal and view consistency.
\subsection{Ablation Studies}

As shown in Fig.~\ref{fig4}, both HAM and 3D-AAG effectively improve multi-view consistency, with their combination achieving the best performance. For generation tasks, as shown in Tab.~\ref{tab2}, the combination yields the most significant improvements, surpassing baseline methods by 50\% and 39.2\%.
\begin{figure}[t]
\centering
\small
\includegraphics[width=0.95\columnwidth]{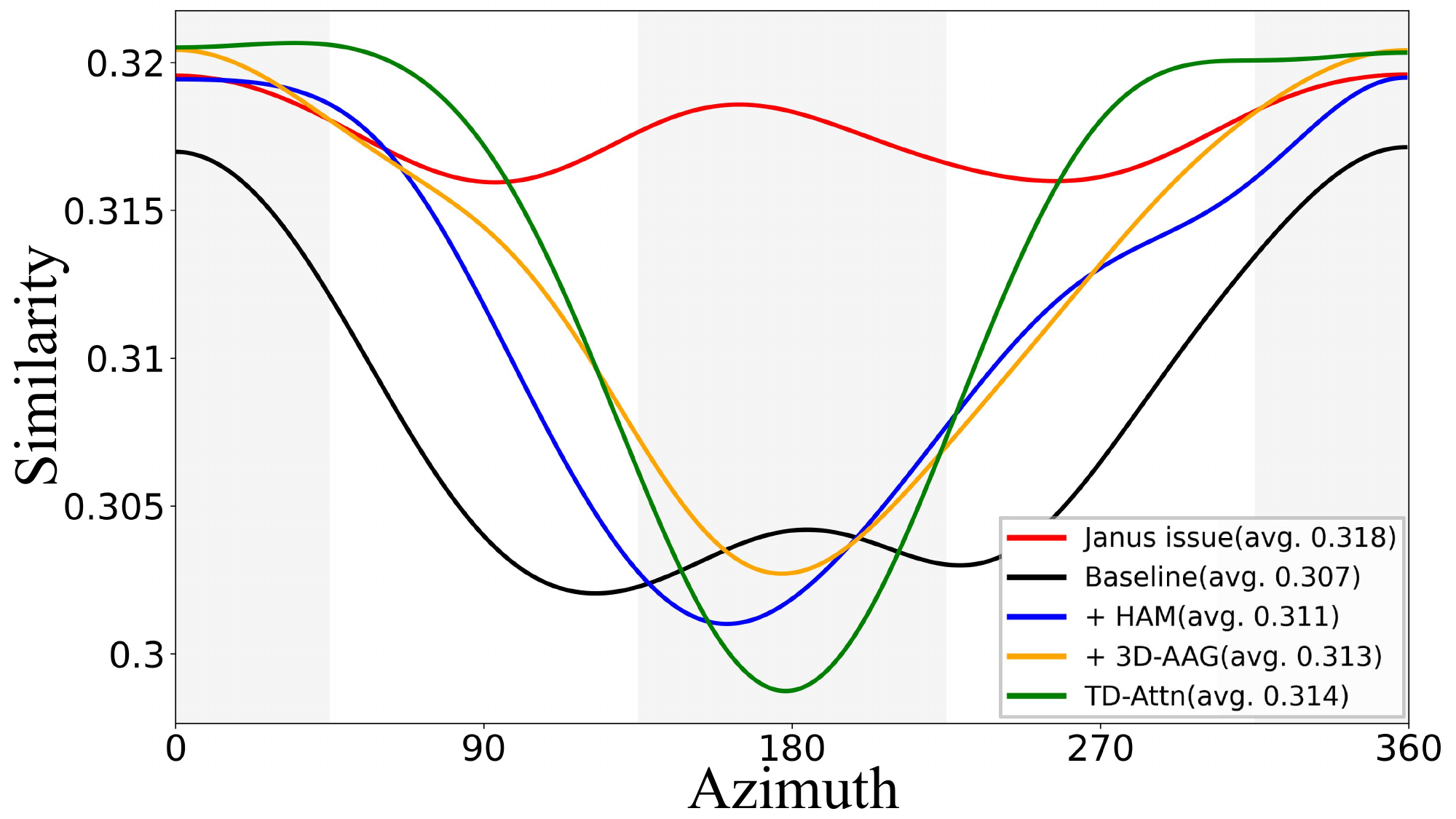} 
\caption{CLIP similarity distribution across different azimuth angles. The plot reveals how multi-view inconsistencies affect similarity scores, where results with the Janus problem exhibit abnormally high scores, while our approach produces a more natural distribution pattern.}
\label{fig5}
\end{figure}

However, we observed that CLIP$_\text{sim}$ metric yields higher scores for generation results with the Janus problem, as shown in Tab.~\ref{tab2}. Fig.~\ref{fig5} presents the CLIP similarity distribution between prompts and rendered results for different methods. Under normal conditions, similarity scores decrease progressively from front view to back view, with back views scoring lowest due to reduced information content. However, when the Janus problem occurs, prior preference view features appear across multiple angles, resulting in abnormally high similarity scores. This observation motivates exploring more suitable evaluation methods. Analyzing similarity score distributions across views proves more effective than relying solely on average similarity scores. Results demonstrate that our method produces similarity distributions more closely aligned with the ideal pattern.

For editing tasks, we employ CLIP$_\text{sim}$ and CLIP$_\text{dir}$ \cite{he2024customize,koo2024posterior,wu2024gaussctrl} as objective evaluation metrics, where CLIP$_\text{dir}$ measures the consistency between image editing direction and text instruction in CLIP feature space. Tab.~\ref{tab3} demonstrates that our method improves the quality of edited 3DGS. The results of the user study for editing tasks show similarly that our method produces more favorable results. Additional HAM applications in T2I models can be found in the appendix.

\section{Conclusion}
We address multi-view inconsistency in 3D tasks by identifying prior view bias as the root cause. Our mathematical analysis reveals that subject-word tokens preferentially activate prior view features, compromising consistency across different perspectives in both generation and editing tasks. TD-Attn provides an effective solution through two key components: a 3D-Aware Attention Guidance Module that constructs view-consistent attention Gaussians by accumulating multi-view cross-attention maps, and a Hierarchical Attention Modulation Module that uses a semantic guidance tree to modulate cross-attention layers for enhanced target semantic awareness. Extensive experiments demonstrate that TD-Attn serves as a universal plugin, significantly reducing inconsistency artifacts in both generation and editing scenarios while maintaining high quality. The framework advances multi-view consistent 3D synthesis and enables fine-grained semantic control across diverse 3D tasks.

\bibliography{main}

\clearpage
\appendix
\twocolumn[
\begin{center}
{\Large \textbf{\textit{Supplementary Materials for}\\
Debiasing Diffusion Priors via 3D Attention for Consistent Gaussian Splatting
}}
\end{center}
\vspace{5em}
]

\maketitle
This supplementary material provides additional experimental validation for the TD-Attn framework. Section A presents a detailed mathematical derivation for view inconsistency, formalizing how anomalous cross-attention (CA) scores accumulate across multi-view optimization and degrade performance when target views deviate from prior distributions. Section B details the implementation specifics of our approach, including complete algorithmic descriptions for both 3D generation (Algorithm 1) and 3D editing (Algorithm 2) pipelines, along with hyperparameter configurations and notation specifications. Section C offers additional experimental analysis that validates the effectiveness of the Hierarchical Attention Modulation module (HAM) in addressing view preference bias and semantic ambiguity.

\section{A. Detailed Mathematical Derivation for View Inconsistency}
The anomalous CA scores between subject-word tokens and prior view features accumulate across the multi-view optimization process in 3D tasks. To mathematically analyze this impact, we need to model the 3D Gaussian Splatting (3DGS) \cite{3dgs} optimization process as a mathematical framework. Previous works \cite{Wang_2023_CVPR} define the probability density function of parameters of 3D volume (e.g., NeRF \cite{mildenhall2021nerf}) as an expectation of the likelihood of 2D rendered images from uniformly sampled object-space viewpoints or a product of conditional likelihoods given a set of uniformly sampled viewpoints and user prompt. Unlike these definitions, our approach defines the density function of the parameters $\phi$ as a product of conditional likelihoods given a set of view-agnostic user prompts $y_{\text{obj}}$ and uniformly sampled view conditions $v^* \in \Lambda$. It can be formulated as:
\begin{equation}
p_{3D}(\phi) = \prod_{v^* \in \Lambda} p_{2D}(z_\phi | y_{\text{obj}}, v^*),
\label{1}
\end{equation}
where $p_{2D}$ is the conditional likelihood of rendered 2D images given 3DGS parameters, and $p_{3D}$ is the joint distribution over 3DGS parameters. Further, applying the logarithm to each side of Eq.~\eqref{1} yields:
\begin{equation}
\begin{aligned}
\log p_{3D}(\phi) &= \log \left(\prod_{v^* \in \Lambda} p_{2D}(z_\phi \mid y_{\text{obj}}, v^*)\right) \\
&= \sum_{v^* \in \Lambda} \log p_{2D}(z_\phi \mid y_{\text{obj}}, v^*),
\end{aligned}
\label{2}
\end{equation}
the gradient of $\phi$ denoted as $\nabla_{\phi} \log p_{3D}(\phi)$, can be expressed as:
\begin{equation}
\begin{aligned}
\nabla_{\phi} \log p_{3D}(\phi) 
&= \sum_{v^* \in \Lambda} \nabla_{\phi} \log p_{2D}(z_\phi \mid y_{\text{obj}}, v^*)\\
&=|\Lambda| \mathbb{E}_{v^* \in \Lambda}[\nabla_{\phi} \log p_{2D}(z_\phi|v^*, y_{\text{obj}})]\\
&= |\Lambda| \mathbb{E}_{v^* \in \Lambda}[\nabla_{z_\phi} \log p_{2D}(z_\phi|v^*, y_{\text{obj}}) \frac{\partial z_\phi}{\partial \phi}],
\end{aligned}
\label{3}
\end{equation}
Eq.~\eqref{3} can be further expanded using Bayes’ theorem:
\begin{equation}
\begin{aligned}
\nabla_{\phi} \log p_{3D}(\phi) 
= &|\Lambda| \mathbb{E}_{v^* \in \Lambda}[(\nabla_{z_\phi} \log p_{2D}(z_\phi) \\&+ \nabla_{z_\phi} \log p_{2D}(v^*, y_{\text{obj}}|z_\phi)) \frac{\partial z_\phi}{\partial \phi}].
\end{aligned}
\label{4}
\end{equation}
To isolate the impact of view bias, we focus on the conditional term by neglecting the unconditional component that primarily affects global structure rather than view-specific features. The second term in Eq.~\eqref{4} can be further expanded as follows:
\begin{equation}
\begin{aligned}
\nabla_{z_\phi} &\log p_{2D}(v^*, y_{\text{obj}}|z_\phi) \\= &\nabla_{z_\phi} \log p_{2D}(v^*|z_\phi)
\\&+ \nabla_{z_\phi} \log p_{2D}(y_{\text{obj}}|z_\phi) \\&+ \nabla_{z_\phi} \log C,
\end{aligned}
\label{5}
\end{equation}
then $C$ can be simplified by using the definition of conditional probability:
\begin{equation}
\begin{aligned}
C &= \frac{p_{2D}(v^*, y_{\text{obj}}|z_\phi)}{p_{2D}(v^*|z_\phi)p_{2D}(y_{\text{obj}}|z_\phi)}\\
&= \frac{p_{2D}(y_{\text{obj}} \mid z_\phi)\, p_{2D}(v^* \mid y_{\text{obj}}, z_\phi)}{p_{2D}(v^* \mid z_\phi)\, p_{2D}(y_{\text{obj}} \mid z_\phi)} \\
&= \frac{p_{2D}(v^* \mid y_{\text{obj}}, z_\phi)}{p_{2D}(v^* \mid z_\phi)}.
\end{aligned}
\label{6}
\end{equation}
when $v^*$ approaches $v_\text{prior}$, i.e, $C \rightarrow 1$, the term $\nabla_{z_\phi} \log C$ does not significantly interfere with gradient computation. However, when $v^*$ is far from $v_\text{prior}$, we have $p_{2D}(v^*|y_{\text{obj}}, z_\phi) \ll p_{2D}(v^*|z_\phi)$, resulting in $C \rightarrow 0$. Consequently, $\nabla_{z_\phi} \log C \ll 0$, resulting in strong negative gradient effects that degrade the performance of text-driven 3D tasks.
\begin{algorithm*}[t]
\caption{Mainflow: 3D Generation via TD-Attn}
\label{alg:generation}
\begin{algorithmic}[1]
\REQUIRE Text prompt $y$, iterations $\text{ITER}_0$, $\text{ITER}_1$, $\text{ITER}_2$, view condition $v^*$, Semantic Guidance Tree $\text{SGT}$
\ENSURE 3D representation $\phi$
\STATE $\phi_0 \leftarrow \text{Frozen-Pre-trained-3D-Generator}(y)$ \COMMENT{Initialize coarse 3DGS}
\FOR{$\phi_i \leftarrow \phi_0$ to $i=\text{ITER}_0$} \COMMENT{Stage 1: Use HAM to modulate CA scores and accumulate 3D attention Gaussian weights}
    \STATE $X_{\phi_i} \leftarrow \mathcal{G}(\phi_i, v^*)$ \COMMENT{Render 3DGS $\phi_i$ from viewpoint $v^*$}
    \STATE $\tilde{X}_{\phi_i} \leftarrow \text{Add-Noise-UNet}(X_{\phi_i})$ \COMMENT{Add noise for DDIM}
    \STATE $\mathcal{S}_{2D}^{v^*}, \hat{X}_{\phi_i} \leftarrow \text{Denoise-UNet-HAM}(\tilde{X}_{\phi_i}, \text{SGT}, y)$ \COMMENT{UNet with HAM and extract 2D CA maps}
    \STATE $\tilde{\phi}_i \leftarrow \text{Inverse-Mapping}(\mathcal{S}_{2D}^{v^*}, v^*, \phi_i)$ \COMMENT{Inverse mapping of 2D CA maps to 3DGS}
    \STATE $\mathcal{L} \leftarrow \mathcal{L}_{\text{ISM}}(X_{\phi_i}, \hat{X}_{\phi_i})$
    \STATE $\phi_{i+1} \leftarrow \text{Update}(\mathcal{L})$ \COMMENT{Update $\phi_i$ with loss function}
\ENDFOR
\FOR{$\phi_j \leftarrow \phi_{\text{ITER}_0}$ to $j=\text{ITER}_1$} \COMMENT{Stage 2: Apply HAM and 3D-AAG jointly}
    \STATE $\widetilde{\mathcal{S}}_{2D}^{v^*}, X_{\phi_j} \leftarrow \mathcal{G}(\phi_j, v^*)$ \COMMENT{Render 3DGS $\phi_j$ from viewpoint $v^*$ to obtain CA maps $\mathcal{S}_{2D}^{v^*}$}
    \STATE $\tilde{X}_{\phi_j} \leftarrow \text{Add-Noise-UNet}(X_{\phi_j})$
    \STATE $\mathcal{S}_{2D}^{v^*}, \hat{X}_{\phi_j} \leftarrow \text{Denoise-UNet-HAM}(\tilde{X}_{\phi_j}, \text{SGT}, y)$
    \STATE $\tilde{\phi}_j \leftarrow \text{Inverse-Mapping}(\mathcal{S}_{2D}^{v^*}, v^*, \phi_j)$
    \STATE $\mathcal{L}_{\text{attn}} = \text{KL}(\text{Softmax}(\widetilde{\mathcal{S}}_{2D}^{v^*}) \parallel \mathcal{I}(\mathcal{S}_{2D}^{v^*}))$ 
    \STATE $\mathcal{L}\leftarrow \mathcal{L}_{\text{ISM}}(X_{\phi_j}, \hat{X}_{\phi_j}) + \lambda_1 \mathcal{L}_{\text{attn}}$
    \STATE $\phi_{j+1} \leftarrow \text{Update}(\mathcal{L})$
\ENDFOR
\FOR{$\phi_k \leftarrow \phi_{\text{ITER}_1}$ to $k=\text{ITER}_2$} \COMMENT{Apply 3D-AAG with standard UNet (no HAM)}
    \STATE $\widetilde{\mathcal{S}}_{2D}^{v^*}, X_{\phi_k} \leftarrow \mathcal{G}(\phi_k, v^*)$
    \STATE $\tilde{X}_{\phi_k} \leftarrow \text{Add-Noise-UNet}(X_{\phi_k})$
    \STATE $\mathcal{S}_{2D}^{v^*}, \hat{X}_{\phi_k} \leftarrow \text{Denoise-UNet}(\tilde{X}_{\phi_k}, y)$ \COMMENT{Standard UNet without HAM}
    \STATE $\tilde{\phi}_k \leftarrow \text{Inverse-Mapping}(\mathcal{S}_{2D}^{v^*}, v^*, \phi_k)$
    \STATE $\mathcal{L}_{\text{attn}} = \text{KL}(\text{Softmax}(\widetilde{\mathcal{S}}_{2D}^{v^*}) \parallel \mathcal{I}(\mathcal{S}_{2D}^{v^*}))$
    \STATE $\mathcal{L} \leftarrow \mathcal{L}_{\text{ISM}}(X_{\phi_k}, \hat{X}_{\phi_k}) + \lambda_1 \mathcal{L}_{\text{attn}}$
    \STATE $\phi_{k+1} \leftarrow \text{Update}(\mathcal{L})$
\ENDFOR
\STATE $\phi \leftarrow \phi_{\text{ITER}_2}$
\end{algorithmic}
\end{algorithm*}

\begin{algorithm*}[t]
\caption{Mainflow: 3D Editing via TD-Attn}
\label{alg:editing}
\begin{algorithmic}[1]
\REQUIRE View condition $v^*$, ground truth images $\text{IMG}_{\text{GT}}$, pre-trained 3DGS $\phi$, Semantic Guidance Tree $\text{SGT}$, edit prompt $y_{\text{edit}}$, mask prompt $y_{\text{mask}}$, ImageReward prompt $y_{\text{IR}}$, iterations $\text{ITER}$
\ENSURE Edited 3D representation $\phi_{\text{edit}}$
\STATE $\phi_{\text{edit}} \leftarrow \phi$ \COMMENT{Initialize with pre-trained 3DGS}
\STATE $\mathcal{D} \leftarrow \text{CameraDataset}(\phi, v^*)$ \COMMENT{Create multi-view dataset}

\FOR{$i \leftarrow 1$ to $\text{ITER}$} \COMMENT{Stage 1: Edit all images with HAM}
    \FOR{$\text{cam}_k \in \mathcal{D}$} \COMMENT{Loop over all views}
        \STATE $X_k \leftarrow \mathcal{D}( \text{cam}_k)$ 
        \STATE $\tilde{X}_k \leftarrow \text{Add-Noise-UNet}(X_k)$
        \STATE $\mathcal{S}_{2D}^{v_k}, \hat{X}_k \leftarrow \text{Denoise-UNet-HAM}(\tilde{X}_k, \text{SGT}, y_{\text{edit}})$ \COMMENT{Edit with HAM}
        \STATE $\text{IMG}_{\text{edit}}[k] \leftarrow \hat{X}_k$ \COMMENT{Store edited image}
    \ENDFOR
\ENDFOR

\STATE $\text{ranking} \leftarrow \text{ImageReward}(\text{IMG}_{\text{edit}}, y_{\text{IR}})$ \COMMENT{Filter with ImageReward}
\STATE $\text{top\_images} \leftarrow \text{SelectTopK}(\text{IMG}_{\text{edit}}, \text{ranking})$
\FOR{$\text{cam}_k \in \mathcal{D}$}\COMMENT{Stage 2: Filtering and Multi-view Fusion}
    \STATE $\text{mask}_k \leftarrow \text{LangSAM}(X_k, y_{\text{mask}})$ \COMMENT{Compute mask with SAM}
    \STATE $\text{MF\_image}_k \leftarrow \text{Reproject}(\text{top\_images}, \text{cam}_k)$ \COMMENT{Multi-view fusion}
    \STATE $\text{MF\_image}_k \leftarrow \text{MF\_image}_k \cdot \text{mask}_k + X_k \cdot (1 - \text{mask}_k)$

    \FOR{$j \leftarrow 1$ to $\text{ITER}$}\COMMENT{Apply 3D-AAG with multi-view guidance}
        \STATE $\widetilde{\mathcal{S}}_{2D}^{v_k}, X_k \leftarrow \mathcal{G}(\phi_{\text{edit}}, \text{cam}_k)$
        \STATE $\tilde{X}_k \leftarrow \text{Add-Noise-UNet}(X_k)$
        \STATE $\mathcal{S}_{2D}^{v_k}, \hat{X}_k \leftarrow \text{Denoise-UNet-MFG}(\tilde{X}_k, \text{SGT}, y_{\text{edit}}, \text{MF\_image}_k)$
        \STATE $\tilde{\phi}_k \leftarrow \text{Inverse-Mapping}(\mathcal{S}_{2D}^{v_k}, v_k, \phi_{\text{edit}})$
        \STATE $\mathcal{L}_{\text{attn}} \leftarrow \text{KL}(\text{Softmax}(\widetilde{\mathcal{S}}_{2D}^{v_k}) \parallel \mathcal{I}(\mathcal{S}_{2D}^{v_k}))$
        \STATE $\mathcal{L} \leftarrow \mathcal{L}_{\text{Edit}}(X_k, \hat{X}_k) + \lambda \mathcal{L}_{\text{attn}}$
        \STATE $\phi_{\text{edit}} \leftarrow \text{Update}(\phi_{\text{edit}}, \mathcal{L})$
    \ENDFOR
\ENDFOR
\end{algorithmic}
\end{algorithm*}
\section{B. Experimental Details}
In this section, we provide comprehensive implementation details of TD-Attn framework. Algorithm 1 and Algorithm 2 demonstrate how TD-Attn serves as a universal plugin for various 3D tasks. 

\textbf{Algorithm~\ref{alg:generation}.} 
The generation pipeline first seeds a coarse 3D Gaussian-Splatting representation $\phi_{0}$ with a frozen text-conditioned 3D generator so that geometry and colour roughly match the prompt $y$. \emph{Stage 1} ($i=0\!\to\!\text{ITER}_{0}$) repeatedly renders $\phi_{i}$ from the target view $v^{*}$, perturbs the image with a DDIM noise step, and denoises it with a HAM-augmented UNet whose CA is steered by the Semantic Guidance Tree (SGT). The subject-word 2D attention maps produced at each step are inversely projected back onto the Gaussians, \emph{accumulating} view-consistent 3D attention weights; optimisation at this stage is driven by $\mathcal{L}_{\text{ISM}}$ \cite{liang2024luciddreamer}. \emph{Stage 2} ($j=\text{ITER}_{0}\!\to\!\text{ITER}_{1}$) activates the full TD-Attn mechanism, a KL divergence between that rendered Gaussian and the denoiser’s 2D CA maps enforces agreement, while HAM continues to refine view-relevant heads, and the total loss becomes $\mathcal{L}_{\text{ISM}}+\lambda_{1}\mathcal{L}_{\text{attn}}$. \emph{Stage 3} ($k=\text{ITER}_{1}\!\to\!\text{ITER}_{2}$) switches off HAM to stabilise fine detail, keeps the 3D/2D attention alignment active, and converges to $\phi_{\text{ITER}_{2}}$, a prompt-faithful and view-consistent 3D asset $\phi$. Throughout all experiments we fix the hyper-parameters to $\text{ITER}_{0}\!=\!200$, $\text{ITER}_{1}\!=\!2000$, $\text{ITER}_{2}\!=\!4000$, and set the attention-balance coefficient to $\lambda_{1}\!=\!10$.

\textbf{Algorithm~\ref{alg:editing}.} 
The editing routine starts from an existing 3DGS~$\phi$ and constructs a multi-view camera set~$\mathcal{D}$ around the object. In \emph{Stage 1} it iterates over all chosen views, adds diffusion noise to each image, and performs a HAM-enhanced denoising step guided by the edit prompt~$y_{\text{edit}}$, thereby producing a set of tentatively edited images~$\text{IMG}_{\text{edit}}$. These candidates are ranked with ImageReward under prompt~$y_{\text{IR}}$, and the best subset is kept. \emph{Stage 2} fuses the accepted edits back into every view: for each camera viewpoint, we employ a language-conditioned Segment Anything Model (SAM) mask~\cite{kirillov2023segment} derived from $y_{\text{mask}}$ to precisely isolate the editable region. Subsequently, the top-ranked images are re-projected into the corresponding view, and through masked compositing operations, we obtain a target-specific guidance image $\text{MF\_image}_k$. A nested optimisation loop (\(j = 1 \rightarrow \text{ITER}\)) performs denoising that is jointly conditioned on the real image, the fused guidance image, and the textual edit, thereby enforcing multi-view, 3D-aware guidance. Each pass inversely projects the refined CA maps to update a 3D attention Gaussian, penalises deviations via the KL loss $\mathcal{L}_{\text{attn}}$, and blends this term with the edit-reconstruction loss $\mathcal{L}_{\text{Edit}}$ to adjust Gaussian positions, opacities, and colours. Repeating the procedure across all cameras aligns view-specific edits into a coherent global change, producing a final $\phi_{\text{edit}}$ that preserves the original geometry while realising the requested, semantically precise modification consistently from every viewpoint. Throughout all experiments we fix the hyper-parameters to $\lambda_{2}\!=\!10$.

\begin{figure}[t]
\centering
\small
\includegraphics[width=1\columnwidth]{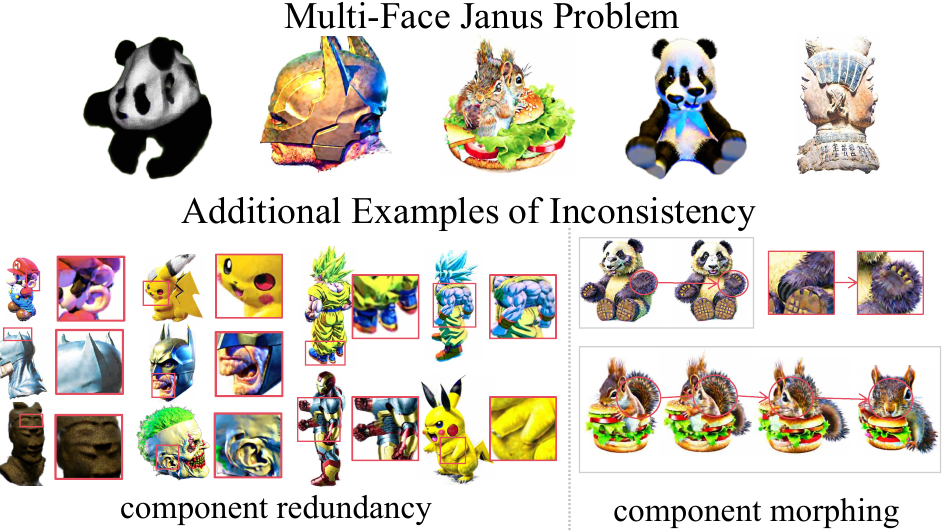} 
\caption{Examples of inconsistencies in generated 3D content.}
\label{PPPfig0}
\end{figure}

\textbf{User Study.} To provide comprehensive subjective evaluation, we designed a user study with 100 participants who evaluated our method against state-of-the-art approaches. Participants rated each generated result on a 10-point scale across two critical dimensions: ``Generation Quality" and ``3D Consistency." Our evaluation protocol ensured objectivity by employing completely randomized sample selection without any preferential curation of results.
In our assessment framework, we established a rigorous definition of view inconsistency that extends beyond the conventional multi-face (Janus) problem. Our evaluation criteria encompass various forms of geometric and semantic inconsistencies as shown in Fig.~\ref{PPPfig0}. 

\textbf{Notation.} 
We provide a comprehensive notation summary in Table~\ref{PPPtab2} to facilitate the understanding of our TD-Attn framework. The table includes mathematical symbols used throughout the paper, covering 3DGS Fundamentals, Text-to-Image Diffusion Model Parameters, View Preference Modeling, 3D-Aware Attention Guidance Module components, and Hierarchical Attention Modulation Module parameters. This standardized notation supports the theoretical derivations and algorithmic descriptions presented in previous sections.

\section{C. Additional Experiments}
We conducted additional experiments to validate the effectiveness of HAM in different scenarios. the first subsection evaluates HAM's view control capabilities in generation tasks, while the second subsection demonstrates its semantic enhancement abilities in editing applications.

\textbf{View-Specific Generation via HAM.} 
We conducted additional experiments to evaluate the performance of HAM in generation tasks, specifically focusing on its ability to overcome the inherent view preference bias in T2I models. Table~\ref{PPPtab1} presents a comparative analysis between standard Stable Diffusion (v2.1 base) \cite{rombach2022high} and Stable Diffusion integrated with our HAM method on view-specific generation tasks. The results demonstrate that for side view generation, HAM integration improves the success rate from 68.6\% to 79.5\%, representing a 10.9 percentage point increase. More remarkably, for back view generation, the standard Stable Diffusion achieves only a 32.4\% success rate, whereas HAM integration significantly elevates this to 75.2\%, constituting a dramatic 42.8 percentage point improvement.

\begin{figure*}[t]

\centering
\includegraphics[width=1\textwidth]{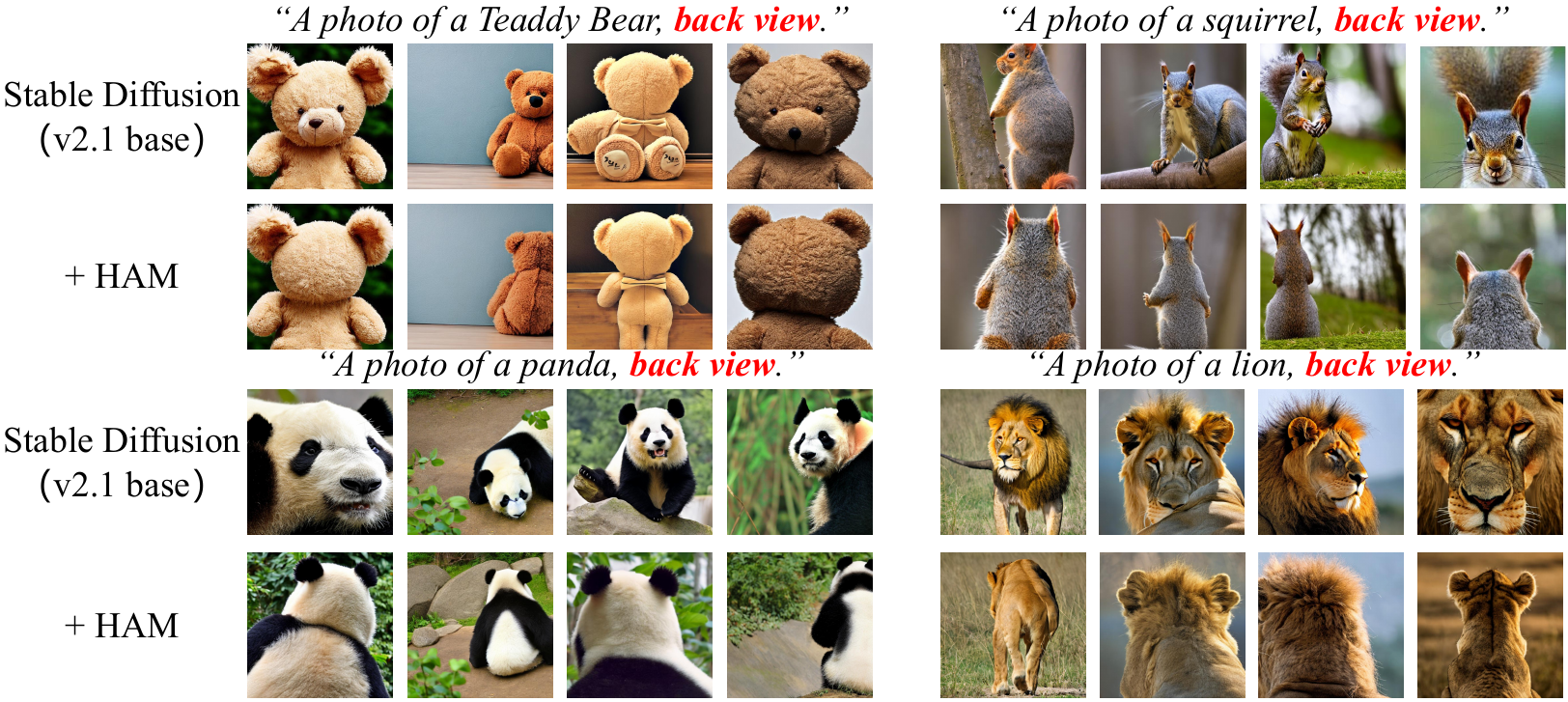} 
\caption{Qualitative comparison of back view generation between Stable Diffusion (v2.1 base) and Stable Diffusion integrated with HAM.}
\label{PPPfig1}
\end{figure*}

Fig.~\ref{PPPfig1} shows qualitative results. Given the open-ended nature of the generation task, we adopt a strict criterion for success: images containing hallucinations (e.g., artifacts or semantic errors) are excluded, and in cases with multiple entities, a single mismatch in viewpoint leads to failure.

\begin{figure*}[t]
\small
\centering
\includegraphics[width=1\textwidth]{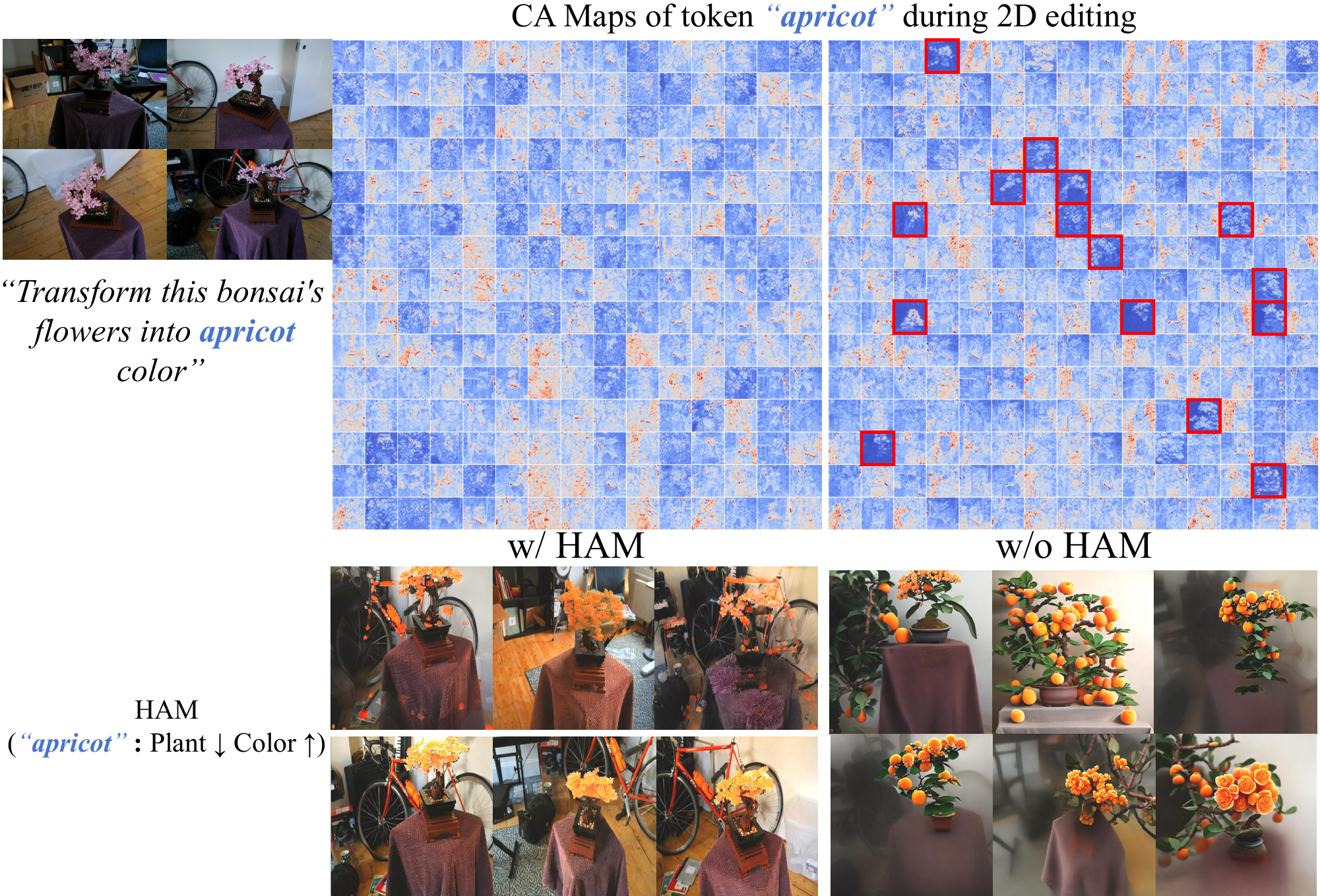} 
\caption{HAM (left) suppresses irrelevant plant activations and correctly attends to color semantics.}
\label{PPPfig2}
\end{figure*}

\begin{table}[t]
\centering
\small
\begin{tabular}{lcc}
\toprule
\multirow{2}{*}{\textbf{Methods}} & \multicolumn{2}{c}{\textbf{Successful Generation Rate}} \\
\cmidrule(lr){2-3}
& \textbf{side view(\%)} $\uparrow$& \textbf{back view(\%)} $\uparrow$\\
\midrule
Stable Diffusion & 68.6 & 32.4 \\
Stable Diffusion+HAM &\textbf{ 79.5} &\textbf{ 75.2} \\
\bottomrule
\end{tabular}
\caption{Successful generation rate comparison.}
\label{PPPtab1}
\end{table}
\textbf{Analysis of HAM in 3D Editing.} 
Figure~\ref{PPPfig2} illustrates a detailed analysis of how HAM addresses semantic ambiguity during editing. When processing prompts such as ``Transform this bonsai's flowers into apricot color," we observe that baseline methods frequently misinterpret the word ``apricot" despite the contextual cue ``color" that should disambiguate its meaning. This phenomenon occurs because diffusion models learn strong correlations between certain words and their most common visual representations during training. The top portion of Figure~\ref{PPPfig2} visualizes the CA maps for the token ``apricot" during the 2D editing process. In the baseline method (right side) \cite{lee2025editsplat}, numerous erroneous activations (highlighted with red boxes) appear in plant-related regions rather than focusing on color-related features. These misplaced attention patterns indicate that the model incorrectly associates ``apricot" with its botanical entity representation despite the explicit contextual qualifier ``color." HAM module addresses this issue by explicitly modeling the semantic structure of concepts through the SGT. For the token ``apricot," HAM identifies competing semantic interpretations (color versus plant) and systematically adjusts attention weights to emphasize the contextually appropriate color semantics while suppressing the irrelevant plant semantics. This intervention is visually represented in the left side of the CA maps, where attention is more uniformly distributed across the flowers without erroneous focal points. The bottom portion of Figure~\ref{PPPfig2} presents the qualitative results of this semantic disambiguation. Without HAM (right side), the editing process materializes ``apricot" as actual apricot fruits, completely misinterpreting the user's intent to change flower color. The results show bonsai plants transformed into fruit-bearing trees with apricots appearing throughout the structure.


\begin{table*}[t]
\centering
\small
\begin{tabular}{lp{6.5cm}}
\toprule
\textbf{Symbol} & \textbf{Meaning} \\
\midrule
$\mu$ & 3D position of Gaussian $\in \mathbb{R}^3$ \\
$\boldsymbol{\Sigma}$ & Covariance matrix of Gaussian $\in \mathbb{R}^{3 \times 3}$ \\
$\boldsymbol{R}$ & Rotation component of factorized covariance matrix \\
$\boldsymbol{S}$ & Scaling component of factorized covariance matrix \\
$\eta$ & Opacity of Gaussian \\
$c$ & View-dependent color (spherical harmonics coefficients) \\
$\phi$ & 3D representation parameters \\
$\theta$ & Parameters of T2I diffusion model \\
$Y$ \& $y$ & Text prompt and its instance \\
$y_{\text{obj}}$ & Object description component in prompt \\
$v^*$ & Target view condition \\
$v_{\text{prior}}$ & Prior view (biased view in training data) \\
$v_{\text{other}}$ & Other views \\
$\mathcal{D}$ & Training dataset \\
$\mathcal{N}$ & Size of training dataset \\
$\epsilon$ & Bias coefficient in view distribution \\
$\mathcal{R}$ & Probability ratio between prior and target views \\
$\Lambda$ & Set of camera views in 3D optimization \\
$z_\phi$ & Rendered 2D image from 3D representation $\phi$ \\
$Q_v$ & Query matrix from noisy image at view $v$ \\
$K_{\text{sbj}}$ & Key matrix of subject-word token \\
$\mathcal{S}_{2D}^v$ & 2D cross-attention map at view $v$ \\
$\widetilde{\mathcal{S}}_{2D}^v$ & Rendered attention map from 3D attention Gaussian \\
$\mathcal{I}(\cdot)$ & Bilinear interpolation function \\
$o_i(p)$ & Opacity of Gaussian $i$ at pixel $p$ \\
$T_i^v(p)$ & Transmittance from pixel $p$ to Gaussian $i$ at view $v$ \\
$w_i$ & Accumulated attention weight for Gaussian $i$ \\
$\mathcal{L}_{\text{attn}}$ & Attention loss \\
$M$ & Number of Semantic-Classes in SGT \\
$F$ & Number of Sub-Classes in SGT \\
$H$ & Number of cross-attention heads \\
$L$ & Number of UNet layers \\
$W_h^f$ & Relevance score of subclass $f$ in head $h$ \\
$W_l^m$ & Relevance score of class $m$ in layer $l$ \\
$W_L^M$ & Layer-wise weights matrix \\
$W_H^F$ & Head-wise weights matrix \\
$\mathcal{U}_m$ & Set of instances in Semantic-Class $m$ \\
$H_l$ & Set of attention heads in layer $l$ \\
$\hat{\mathcal{A}}_h$ & Modulated attention computation at head $h$ \\
$\lambda$ & Modulation coefficient \\
$\lambda_1, \lambda_2$ & Attention balance coefficients \\
$\mathcal{L}_{\text{Gen}}$ & Generation task loss \\
$\mathcal{L}_{\text{Edit}}$ & Editing task loss \\
f$_{\text{mf}}$ & Frequency of multi-face (Janus) problem \\
f$_{\text{inc}}$ & Frequency of view inconsistency \\
\bottomrule
\end{tabular}
\caption{Notation summary for TD-Attn. The table lists mathematical symbols used throughout the paper with their corresponding meanings.}
\label{PPPtab2}
\end{table*}

\end{document}